\documentclass[11pt,a4paper]{article}
\usepackage[hyperref]{acl2021}
\usepackage{times}
\usepackage{latexsym}

\usepackage{microtype}
\usepackage{amsmath}
\usepackage{amsfonts}
\usepackage{algorithm}
\usepackage[noend]{algpseudocode}
\usepackage{bm}
\usepackage{bbm}
\usepackage{booktabs,colortbl}
\usepackage{balance}
\usepackage{color}
\usepackage{CJKutf8}
\usepackage{graphicx} 
\usepackage{graphics}
\usepackage{multirow}
\usepackage{paralist}
\usepackage{subfig}
\usepackage{textcomp}
\usepackage[normalem]{ulem}
\usepackage{arydshln}
\usepackage{enumitem}
\usepackage{CJK}
\usepackage{array}
\usepackage{pifont}
\usepackage{ascii}

\aclfinalcopy 
\def\aclpaperid{3892} 

\definecolor{halfred}{RGB}{255, 128, 128}

\definecolor{zptu}{RGB}{18, 141, 21}

\definecolor{shuo}{RGB}{250, 0, 0}

\title{On the Language Coverage Bias for Neural Machine Translation}

\author{
Shuo Wang$^1$ \  Zhaopeng Tu$^2$ \  Zhixing Tan$^1$ \ 
Shuming Shi$^2$ \  Maosong Sun$^{1,3}$ \   Yang Liu$^{1,3,4}$\\
$^1$Dept. of Comp. Sci. \& Tech., Institute for AI, BNRist Center, Tsinghua University\\
$^3$Beijing Academy of Artificial Intelligence \ $^4$Institute for AIR, Tsinghua University\\
$^1${\href{mailto:wangshuo.thu@gmail.com}{\asciifamily{wangshuo.thu@gamil.com}}} $^1${\href{mailto:liuyang2011@tsinghua.edu.cn}{\asciifamily{\{zxtan, sms, liuyang2011\}@tsinghua.edu.cn}}}\\
$^2$Tencent AI Lab\\
$^2${\href{mailto:zptu@tencent.com}{\asciifamily{\{zptu, shumingshi\}@tencent.com}}}\\
}

\date{}

\begin{document}
\maketitle
\begin{abstract}
Language coverage bias, which indicates the content-dependent differences between sentence pairs originating from the source and target languages, is important for neural machine translation (NMT) because the target-original training data is not well exploited in current practice. By carefully designing experiments, we provide comprehensive analyses of the language coverage bias in the training data, and find that using only the source-original data achieves comparable performance with using full training data. Based on these observations, we further propose two simple and effective approaches to alleviate the language coverage bias problem through explicitly distinguishing between the source- and target-original training data, which consistently improve the performance over strong baselines on six WMT20 translation tasks. Complementary to the translationese effect, language coverage bias provides another explanation for the performance drop caused by back-translation~\cite{Marie:2020:ACL}. We also apply our approach to both back- and forward-translation and find that mitigating the language coverage bias can improve the performance of both the two representative data augmentation methods and their tagged variants~\cite{Caswell:2019:WMT}.

\end{abstract}

\section{Introduction}

\begin{figure}[t]
    \centering
    \subfloat[English-Original]{
    \includegraphics[height=0.2\textwidth]{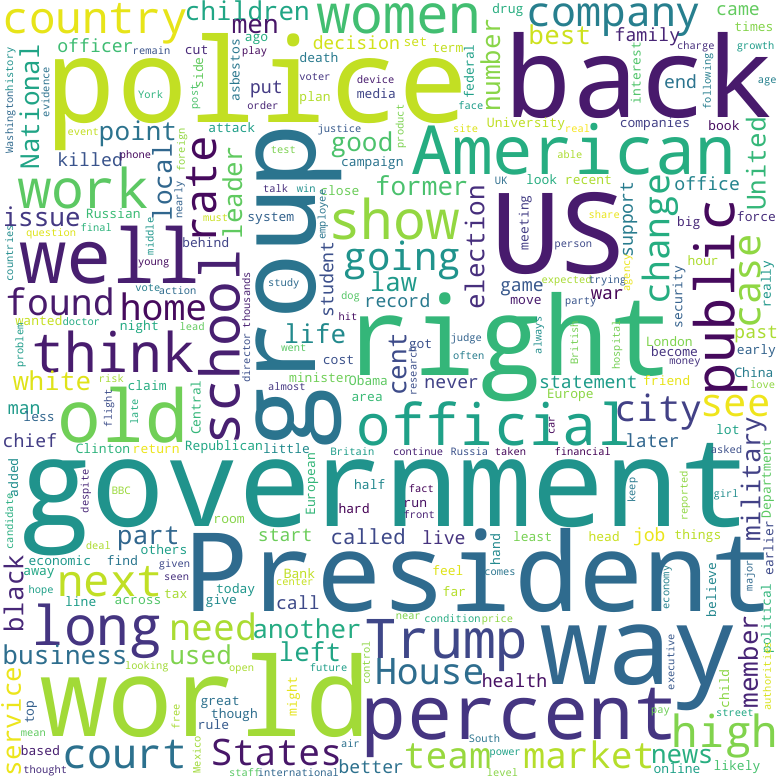}}
    \hspace{0.03\textwidth}
    \subfloat[German-Original]{
    \includegraphics[height=0.2\textwidth]{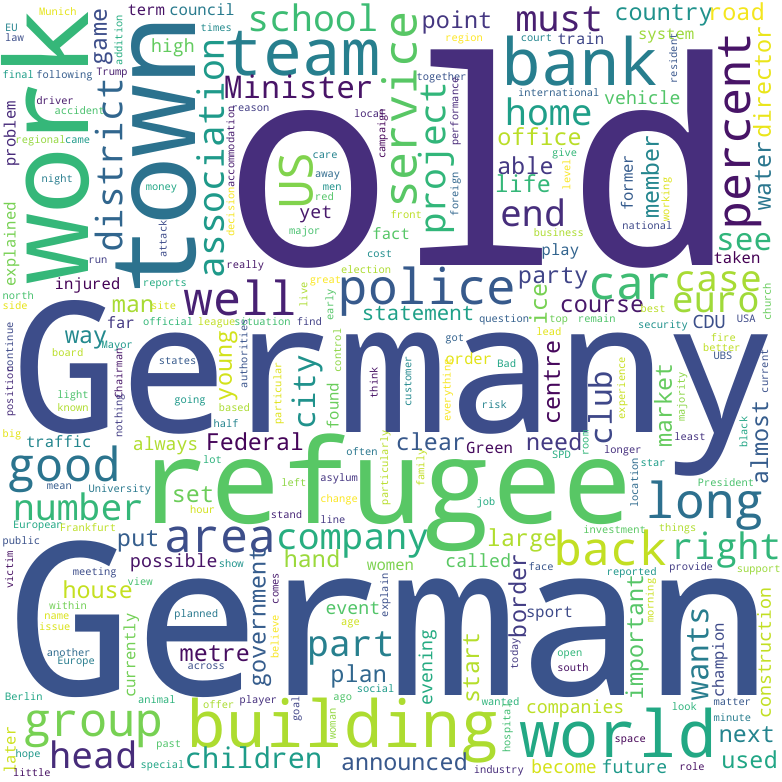}}
\caption{Example of language coverage bias illustrated by word clouds that are plotted at the English side of sentence pairs in the En-De test sets from WMT10 to WMT18. The test sets consist of English-original and German-original sentence pairs.
}
\label{fig:language-bias}
\end{figure}

In recent years, there has been a growing interest in investigating the effect of original languages in parallel data on neural machine translation~\cite{Barrault:2020:WMT,Edunov:2020:ACL,Marie:2020:ACL}.
Several studies have shown that target-original test examples\footnote{Target-original test examples are sentence pairs that are translated from the target language into the source language.} can lead to distortions in automatic and human evaluations, which should be omitted from machine translation test sets~\cite{Barrault:2019:WMT,Zhang:2019:WMT,Graham:2020:EMNLP}.
Another branch of studies report that target-original test data leads to discrepant conclusions: back-translation only benefits the translation of target-original test data while harms that of source-original test data~\cite{Edunov:2020:ACL,Marie:2020:ACL}. 
They attribute these phenomena to the reason that human-translated texts (i.e., {\em translationese}) exhibit formal and stylistic differences that set them apart from the texts originally written in that language~\cite{Baker:1993:TT,Volansky:2015:Translationese,Zhang:2019:WMT}.

Complementary to the translationese bias, which is {\em content-independent}~\cite{Volansky:2015:Translationese},
we identify another important problem, namely {\em language coverage bias}, which refers to the {\em content-dependent} differences in data originating from different languages.
These differences stem from the diversity of regions and cultures. While the degree of the translationese bias varies across different translators~\cite{Toral:2019:post}, language coverage bias is an intrinsic bias between the source- and target-original data, which is hardly affected by the ability of the translator.
Figure~\ref{fig:language-bias} shows an example, where the contents in English- and German-original texts differ significantly due to language coverage bias.

To investigate the effect of language coverage bias in the training data on NMT models,
we propose an automatic method to identify the original language of each training example, which is generally unknown in practical corpora. Experimental results on three large-scale translation corpora show that there exists a significant performance gap between NMT models trained on the source- and target-original data, which have different vocabulary distributions, especially for content words.
Since the target-original training data performs poorly in translating content words, using only the source-original data achieves comparable performance with using full training data. These findings motivate us to explore other data utilization methods rather than indiscriminately mixing the source- and target-original training data.

We propose to alleviate the language coverage bias problem by explicitly distinguishing between the source- and target-original training data.
Specifically, two simple and effective methods are employed: {\em bias-tagging}
and {\em fine-tuning}.
Experimental results show that both approaches consistently improve the performance on six WMT20 translation tasks.
Language coverage bias also provides another explanation for the failure of back-translation on the source-original test data, complementary to the translationese effect~\cite{Marie:2020:ACL}.
We further validate our approach in the monolingual data augmentation scenario, where the language coverage bias problem would be more severe due to the newly introduced monolingual data. 

\paragraph{Contributions} The main contributions of our work are listed as follows:
\begin{itemize}
    \item We demonstrate the necessity of studying the language coverage bias for NMT, and identify that using the target-original data can cause poor translation adequacy on content words.
    
    \item We address the language coverage bias induced by the target-original data by explicitly distinguishing the original languages, which can significantly improve the translation performance on six WMT20 translation tasks.
    
    \item We show that alleviating the language coverage bias also benefits monolingual data augmentation, which can improve both back- and forward-translation and their tagged variants~\cite{Caswell:2019:WMT}.
\end{itemize}

\section{Experimental Setup}
\label{sec:setup}
\paragraph{Data}
We conducted experiments on six WMT20 benchmarks~\cite{Barrault:2020:WMT}, including English$\Leftrightarrow$German (En$\Leftrightarrow$De), English$\Leftrightarrow$Chinese (En$\Leftrightarrow$Zh), and English$\Leftrightarrow$Japanese (En$\Leftrightarrow$Ja) news translation tasks.
The preprocessed training corpora contain 41.0M, 21.8M, and 13.0M sentence pairs for En$\Leftrightarrow$De, En$\Leftrightarrow$Zh, and En$\Leftrightarrow$Ja, respectively.
We used the monolingual data that is publicly available in WMT20 to train the proposed original language detection model (Section~\ref{sec:detection}) and data augmentation (Section~\ref{sec:data-augmentation}).
The Appendix lists details about the data preprocessing.

For En$\Leftrightarrow$De and En$\Leftrightarrow$Zh, 
we used {newstest2019} as the validation sets.
For En$\Leftrightarrow$Ja, we split the official validation set released by WMT20 into two parts by the original language and only used the corresponding part for each direction. 
We used {newstest2020} as the test sets for all the six tasks.
We reported the Sacre BLEU~\cite{post2018call}, as recommended by WMT20.

\paragraph{Model}
We used the Transformer-Big ~\cite{Vaswani:2017:NeurIPS} model, which consists of a 6-layer encoder and a 6-layer decoder, and the hidden size is 1024.
Recent studies showed that training on large batches can further boost model performance~\cite{Ott:2018:WMT,Wu:2018:ICLR}. Accordingly, we followed their settings to train models with batches of approximately 460k tokens. Please refer to the Appendix for more details about model training.
We followed \citet{Ng:2019:WMT} to use the Transformer-Big decoder as our language models, which are used to detect the original language and measure translation fluency. Language models are also trained with large batches~\cite{Ott:2018:WMT}.

\section{Observing Language Coverage Bias}
\label{sec:obs-languagecoveragebias}

In this study, we first establish the existence of language coverage bias (Section~\ref{sec:existence}), and show how the bias affects NMT performance (Section~\ref{sec:effect}). To this end, we propose an automatic method to detect the original language of each training example (Section~\ref{sec:detection}), which is often not available in large-scale parallel corpora~\cite{Riley:2020:ACL}.

\subsection{Detecting Original Languages}
\label{sec:detection}

\paragraph{Detection Method}
Intuitively, we use a large-scale monolingual dataset to estimate the distribution of the contents covered by each language.
For each training example, we compare its similarities to the distributions of source and target languages, based on which we determine its original language.

Formally, let $\mathcal{D}_{s}$ and $\mathcal{D}_{t}$ denote the source-side and target-side distributions of the covered contents. Given a training example $\langle \mathbf{x}, \mathbf{y} \rangle$, the probability that it is covered by one language (represented as $\mathcal{D}_{s}$ and $\mathcal{D}_{t}$) can be expressed as 
\begin{equation*}
\label{eq:bayes}
    \begin{aligned}
    P(\mathcal{D}_{s} | \langle \mathbf{x}, \mathbf{y} \rangle) & = \frac{P(\mathcal{D}_{s}) P(\langle \mathbf{x}, \mathbf{y} \rangle | \mathcal{D}_{s})}{P(\langle \mathbf{x}, \mathbf{y} \rangle)}, \\
    P(\mathcal{D}_{t} | \langle \mathbf{x}, \mathbf{y} \rangle) & = \frac{P(\mathcal{D}_{t}) P(\langle \mathbf{x}, \mathbf{y} \rangle | \mathcal{D}_{t})}{P(\langle \mathbf{x}, \mathbf{y} \rangle)}.
    \end{aligned}
\end{equation*}
We use a score function to denote the difference between the two probabilities:
\begin{equation*}
\label{eq:score-func}
\begin{split}
\text{\em score} = \log P(\mathcal{D}_{s} | \langle \mathbf{x}, \mathbf{y} \rangle) - \log P(\mathcal{D}_{t} | \langle \mathbf{x}, \mathbf{y} \rangle),\\
= \log P(\langle \mathbf{x}, \mathbf{y} \rangle | \mathcal{D}_{s}) - \log P(\langle \mathbf{x}, \mathbf{y} \rangle | \mathcal{D}_{t}) + c,
\end{split}
\end{equation*}
where $c=\log P(\mathcal{D}_{s}) - \log P(\mathcal{D}_{t})$, which has a constant value when the source and target monolingual datasets are given. 
Intuitively, examples with higher score values are more likely to be source-original while those with lower score values are more likely to be target-original data. We train language models $\bm{\theta}^{lm}_{s}$ and $\bm{\theta}^{lm}_{t}$ on the source- and target-language monolingual data
to estimate the conditional probabilities:
\begin{equation*}
\label{eq:score-lm}
\begin{aligned}
P(\langle \mathbf{x}, \mathbf{y} \rangle | \mathcal{D}_{s}) &= P( \mathbf{x} | \bm{\theta}^{lm}_{s}), \\
P(\langle \mathbf{x}, \mathbf{y} \rangle | \mathcal{D}_{t}) &= P( \mathbf{y} | \bm{\theta}^{lm}_{t}).
\end{aligned}
\end{equation*}
Accordingly, the score can be rewritten as
\begin{equation}
\label{eq:bias-score}
\begin{split}
\text{\em score} = \log P( \mathbf{x} | \bm{\theta}^{lm}_{s}) - \log P( \mathbf{y} | \bm{\theta}^{lm}_{t}) + c.
\end{split}
\end{equation}
We label examples as source-original if their score values are positive, and the other examples as target-original.
To find a specific constant for each language pair, we tune the value of $c$ to obtain the best classification performance on the validation sets, where the original languages are known.

\begin{table}[t]
  \centering
  \begin{tabular}{cccc}
  \toprule
  \bf Method    &   \bf En-Zh   & \bf En-Ja &   \bf En-De   \\
\midrule
  FT    & 83.6 & 83.7 & 86.6 \\
  Ours  & \bf 84.4 & \bf 91.5 & \bf 88.7 \\
  \bottomrule
  \end{tabular}
  \caption{F1 scores of detecting original languages in the test sets. ``FT" denotes the forward translation classifier proposed by \citet{Riley:2020:ACL}. }
  \label{tab: classification}
\end{table}

\paragraph{Detection Accuracy}
We evaluated the detection method on the mixture of the test sets of bidirectional translation tasks in WMT20 for each language pair. For comparison, we re-implemented the CNN-based forward-translation (FT) classifier proposed by \citet{Riley:2020:ACL}. The FT classifier and the language models used in our method were trained on the same monolingual data sets.
Table~\ref{tab: classification} shows that our method outperforms the FT classifier in all language pairs.
In addition, our model also outperforms the FT approach on detecting noisy training data, which leads to an improvement in translation performance (please refer to Table~\ref*{tab:effect-of-detection} in the Appendix for more results).

\subsection{Existence of Language Coverage Bias}
\label{sec:existence}

In this section, we validate the existence of language coverage bias by (1) comparing the performance of NMT models trained on data with different original languages, and (2) directly calculating the divergence between the vocabulary distributions of the source- and target-original data.

\begin{figure}[t]
    \centering
    \subfloat[En$\Rightarrow$Zh]{
        \includegraphics[height=0.26\textwidth]{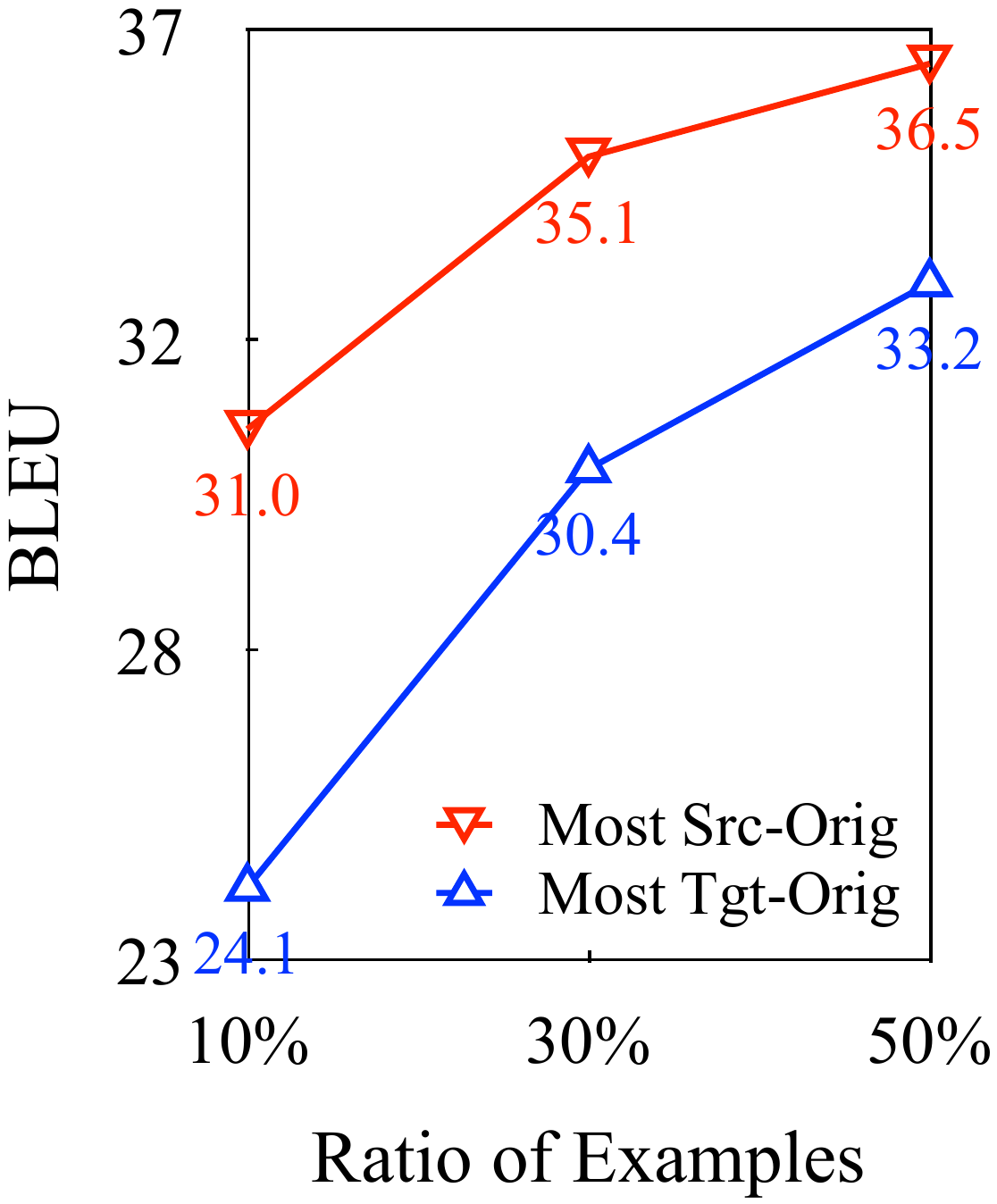}
    }\hfill
    \subfloat[Zh$\Rightarrow$En]{
        \includegraphics[height=0.26\textwidth]{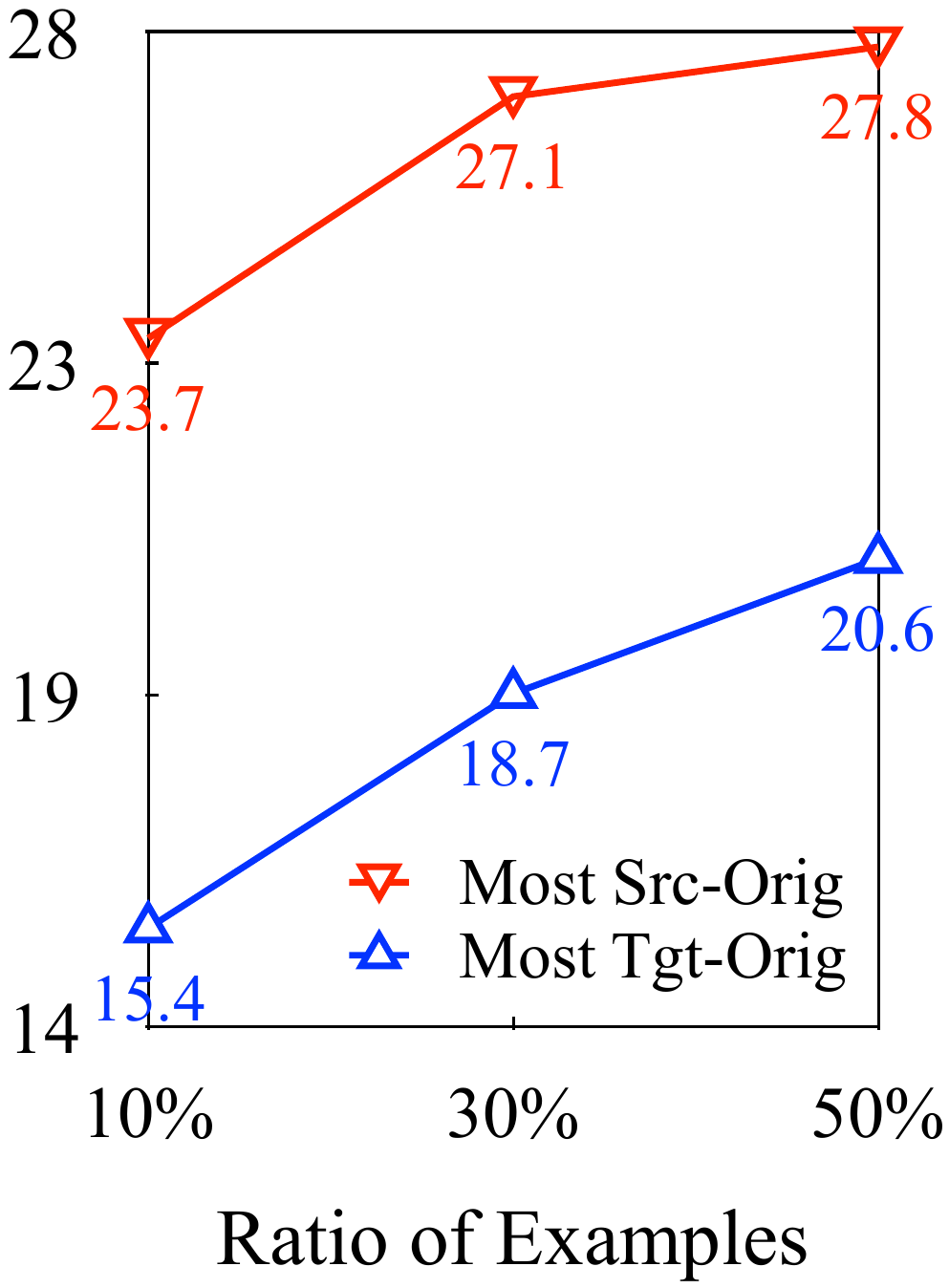}
    }
\caption{Translation performance on the validation sets of the En$\Leftrightarrow$Zh translation task for different ratios of most source- and target-original training examples.}
\label{fig:ratio-translation}
\end{figure}

\paragraph{Translation Performance}
Once all the training examples are assigned a score by the detection method (Eq.~(\ref{eq:bias-score})), we regard $R\%$ of examples with the highest scores as the most source-original examples, and $R\%$ of examples with the least scores as the most target-original examples.
We investigate the effect of $R\%$ on translation performance, as shown in Figure~\ref{fig:ratio-translation}. Clearly, using the most source-original examples significantly outperforms using its target-original counterparts, demonstrating that the source- and target-original data indeed differ greatly from each other.
To rule out the effect of data scale, we treat 50\% of data with the highest scores as source-original data, and the same amount of data with the least scores as target-original data in the following experiments by default.

Since some recent works find that BLEU might be affected by the translationese problem~\cite{Edunov:2020:ACL,Freitag:2020:BLEU}, we have also conducted a side-by-side human-evaluation on the Zh$\Rightarrow$En development set, where 500 randomly sampled examples were evaluated by six persons (agreement~\cite{Fleiss:1971:agree}: Fleiss’ Kappa=0.46). 37.0\% of outputs using the source-original data are better than using target-original data, and 21.0\% are worse. By manually checking the outputs, we find using only the target-original data tends to omit important source contents (e.g., named entities) either by totally ignoring some contents or by using pronouns instead. The human-evaluation shows the same trend with the BLEU score presented in Figure~\ref{fig:ratio-translation}.  Given that conducting human-evaluation on all the six translation tasks is time-consuming and labor-intensive, we use automatic measures to further investigate this problem in Section~\ref{sec:effect}.

\begin{table}[t]
  \centering
  \begin{tabular}{crrr}
  \toprule
  \multirow{2}{*}{\bf Data}  &  \multicolumn{3}{c}{\bf JS ($\times 10^{-5}$)}\\
  \cmidrule(lr){2-4}
  & \em All & \em Content & \em Function\\
  \midrule
  Random & 4 & 10 & 0\\
  S vs T & 745 & 1503 & 261\\
  \bottomrule
  \end{tabular}
  \caption{JS divergence of the vocabulary distributions between the source- and target-original data (``S vs T'') on the training set of WMT20 En$\Leftrightarrow$Zh. ``{\em All}'', ``{\em Content}'', and ``{\em Function}'' denote all words, content words, and function words respectively. For reference, we also report the JS divergence between randomly selected 50\% examples and the others (``Random'').}
  \label{tab:language-bias-bilingual}
\end{table}

\paragraph{Vocabulary Distributions}
Complementary to previous studies that focus on the {\em content-independent} stylistic difference~\cite{Volansky:2015:Translationese} between  translationese and original texts~\cite{Riley:2020:ACL,Edunov:2020:ACL,Marie:2020:ACL}, we investigate the {\em content-dependent} language coverage bias between the source- and target-original data in this experiment. Intuitively, if the language coverage bias exists, the vocabulary distributions of the source- and target-original data should differ greatly from each other, since the covered issues tend to have different frequencies between them~\cite{Dave:2000:JC}.
We use the Jensen-Shannon (JS) divergence~\cite{Lin:1991:TIT} to measure the difference between two vocabulary distributions $p$ and $q$:
\begin{equation}
\begin{split}
 \mathrm{JS}\left (p || q \right ) = \frac{1}{2} \left (
\mathrm{KL} (p || \frac{p+q}{2} ) + \mathrm{KL} (q || \frac{p+q}{2})
\right ), \nonumber
\end{split}
\end{equation}
where $\mathrm{KL}(\cdot||\cdot)$ is the KL divergence~\cite{KL:1951:information} of two distributions.

Table~\ref{tab:language-bias-bilingual} shows the JS divergence of the vocabulary distributions between the source- and target-original data. We also divide the words into content words and functions words based on their POS tags, since content words are more related to the language coverage bias, while the function words are more related to the stylistic and structural differences between the translationese and original texts~\cite{Lembersky:2011:EMNLP,Volansky:2015:Translationese}.
The JS divergence between the source- and target-original data are 186$\times$ larger than that between randomly split data, which is mainly due to the difference between content words.
Results for different ratios $R\%$ and other language pairs can be found in Appendix (Tables~\ref*{tab:language-bias-ratios} and~\ref*{tab:language-bias-lanuages}), where the trend holds in all cases, supporting our claim of the existence of language coverage bias.

\subsection{Effect of Language Coverage Bias}
\label{sec:effect}

In this section, we investigate the effect of language coverage bias on NMT models.
\begin{table}[t]
  \centering
  \scalebox{0.9}{
  \begin{tabular}{c cc cc cc}
  \toprule
  {\bf Data} & \multicolumn{2}{c}{\bf En-Zh}  &  \multicolumn{2}{c}{\bf En-Ja} &  \multicolumn{2}{c}{\bf En-De}\\
  \cmidrule(lr){2-3} \cmidrule(lr){4-5} \cmidrule(lr){6-7}
  \bf Origin &   $\Rightarrow$   &   $\Leftarrow$    &   $\Rightarrow$   &   $\Leftarrow$    &   $\Rightarrow$   &   $\Leftarrow$\\
  \midrule
  Target    & 33.2 & 20.6 & 30.5 & 15.4 & 39.3 & 37.4 \\
  Source    & \underline{36.5} & \bf 27.8 & \bf 35.3 & \underline{17.9} & \underline{41.7} & \bf 42.5 \\
  Both  & \bf 36.6 & \underline{27.5} & \underline{34.9} & \bf 18.5 & \bf 42.3 & \underline{42.2} \\
  \bottomrule
  \end{tabular}}
  \caption{Sacre BLEU of using different sets of training data on validation sets.
  We highlight the {\bf highest} score in bold and the \underline{second-highest} score with underlines.
  }
  \label{tab:effect-translation}
\end{table}

\begin{table}[t]
  \centering
  \scalebox{0.9}{
  \begin{tabular}{c ccc ccc}
  \toprule
  {\bf Data}  &  \multicolumn{3}{c}{\bf En$\Rightarrow$Zh}  &  \multicolumn{3}{c}{\bf En$\Leftarrow$Zh}\\
  \cmidrule(lr){2-4} \cmidrule(lr){5-7}
  \bf Origin & \em noun & \em verb & \em adj  & \em noun & \em verb & \em adj\\
  \midrule
  Target    & 67.6 & 52.0 & 64.3 &  53.8 & 38.0 & 57.0 \\
  Source    & \underline{69.7} & \underline{54.0} & \bf 66.2 & \bf 61.8 & \bf 44.1 & \bf 63.9 \\
  Both  & \bf 69.9 & \bf 54.1 & \underline{65.9} & \underline{61.2} & \underline{43.8} & \underline{63.4} \\
  \bottomrule
  \end{tabular}}
  \caption{Translation adequacy of different types of content words measured by F-measure~\cite{Neubig:2019:compare}. The results are reported on the validation sets.}
  \label{tab:adequacy}
\end{table}

\paragraph{Using only the source-original data achieves comparable performance with using full data.} 
Table~\ref{tab:effect-translation} lists the translation performances of NMT models trained on only the source- or target-original data and on both of them. The results show that using only the source-original data significantly outperforms using the target-original data in all language pairs, which reconfirm the necessity of studying the language coverage bias for NMT.
It should be emphasized that using only the source-original data (i.e. 50\% of the whole training set) achieves translation performances on par with using full training data.
In the following experiments, we investigate why using target-original data together cannot further improve the performance.

\paragraph{Using additional target-original data does not consistently improve translation adequacy.}
To rule out the effect of translationese and focus on the content-dependent difference caused by the language coverage bias, we examine the translation adequacy of content words in Table~\ref{tab:adequacy}~\footnote{We only list the results on En$\Leftrightarrow$Zh due to space limit. Please refer to Table~\ref*{tab:adequacy-other-languages} in the Appendix for the translation quality on other language pairs.}. We follow \citet{Raunak:2020:LongTail} to use F-measure~\cite{Neubig:2019:compare} to quantify the translation accuracy of specific types of words.

Compared with the source-original data, using only the target-original data greatly reduces the translation accuracy of content words, which we attribute to the divergence of the content word distributions between the source- and target-original data.
The results also indicate that indiscriminately using all the training data can not consistently improve the translation adequacy of content words over using only source-original data, and in some cases using all the data is even harmful to the adequacy on content words.
Table~\ref{tab:case} shows an example,
which suggests that using only the target-original data tends to omit content words.
This problem is potentially caused by that some content words at the source-side are less or even not visible in the target-original data, and indiscriminately adding target-original data induces a distribution shift on the content word distribution.

\begin{table}[t]
    \centering
    \scalebox{0.88}{
    \begin{tabular}{cl}
    \toprule
    \bf Input & \begin{CJK}{UTF8}{gbsn}{\color{blue}大闸蟹}是{\color{red}巴城}最为知名的形象代言人。\end{CJK}\\
    \hline
    \multirow{2}{*}{\bf Refer.} & The {\color{blue} hairy crab} is the most famous image\\
    & spokesperson in {\color{red} Bacheng}. \\
    \midrule
    \bf Target & It is one of the city's most well-known \\
    \bf Orig.  & image spokesmen. \\
    \cmidrule{1-2}
    \bf Source & {\color{blue} Hairy crabs} are the most well-known \\
    \bf Orig.  & image spokesmen of {\color{red} Bacheng}. \\
    \cmidrule{1-2}
    \bf Both & It is the best-known icon of {\color{red} Bacheng}. \\
    \bottomrule
    \end{tabular}}
    \caption{An example of the outputs of NMT models trained on different sets of data.
    Using the target-original data tends to omit content words.}
    \label{tab:case}
\end{table}

\paragraph{Using additional target-original data only slightly improves the structural fluency.}
Recently, \citet{Edunov:2020:ACL} claim that using additional back-translated data can improve translation fluency.
Target-original bilingual data is similar to back-translated data since both of them are constructed by translating sentences from the target language into the source language.
One question naturally arises: {\em can target-original bilingual data improve the fluency of NMT models?} 

To answer the above question, we measure the fluency of outputs with language models trained on the monolingual data as described in Section~\ref{sec:setup}.
Previous study finds that different perplexities could be caused by specific contents rather than structural differences~\cite{Lembersky:2011:EMNLP}. Specifically, some source-original contents are of low frequency in the target-language monolingual data (e.g., ``Bacheng" in Table~\ref{tab:case}), thus the language model trained on the target-language monolingual data tends to assign higher perplexities to outputs containing more source-original content words.
To rule out this possibility and check whether the outputs are structurally different, we follow~\citet{Lembersky:2011:EMNLP} to abstract away from the content-specific features of the outputs to measure their fluency at the syntactic level.
Table~\ref{tab:fluency} shows the results.
Although using only the source-original data results in high perplexities measured by vanilla language models, the perplexities of NMT models trained on different data are close to each other at the syntactic level. Using additional target-original data only slightly reduces the perplexity at the syntactic level over using only the source-original data.

\begin{table}[t]
  \centering
  \scalebox{0.9}{
  \begin{tabular}{c lr lr}
  \toprule
  {\bf Data} & \multicolumn{2}{c}{\bf No Abs.} & \multicolumn{2}{c}{\bf Cont. Abs.} \\
  \cmidrule(lr){2-3} \cmidrule(lr){4-5}
  \bf Origin    & \em PPL & \em Diff. & \em PPL & \em Diff. \\
  \midrule
  \multicolumn{5}{c}{\em WMT20 En$\Rightarrow$Zh} \\
  \midrule
  Target    & 38.4 & $-$6.3\% & 14.0 & $-$2.8\% \\
  Source    & 44.0 & $+$7.3\% & 14.6 & $+$1.4\% \\
  \midrule
  Both  & 41.0 & - & 14.4 & - \\
  \midrule
  \multicolumn{5}{c}{\em WMT20 En$\Leftarrow$Zh} \\
  \midrule
  Target    & 25.9 & $-$5.8\% & 13.4 & $-$3.6\% \\
  Source    & 31.0 & $+$12.7\% & 14.2 & $+$2.2\% \\
  \midrule
  Both  & 27.5 & - & 13.9 & - \\
  \bottomrule
  \end{tabular}}
  \caption{Translation fluency measured by the perplexities (i.e., PPL) of language models with different levels of lexical abstraction, ``Diff." means the relative change with respect to ``Both". ``No Abs.'' denotes no abstraction (i.e., vanilla LM), ``Cont. Abs.'' denotes abstracting all content words with their corresponding POS tags. The results are reported on the validation sets.}
  \label{tab:fluency}
\end{table}

\section{Addressing Language Coverage Bias}

\begin{table*}[t]
  \centering
  \begin{tabular}{l ll ll ll c}
  \toprule
  \multirow{2}{*}{\bf Method} & \multicolumn{2}{c}{\bf En-Zh}  &  \multicolumn{2}{c}{\bf En-Ja} &  \multicolumn{2}{c}{\bf En-De} &  \multirow{2}{*}{\bf Average} \\
  \cmidrule(lr){2-3} \cmidrule(lr){4-5} \cmidrule(lr){6-7}
  & \multicolumn{1}{c}{$\Rightarrow$} & \multicolumn{1}{c}{$\Leftarrow$}
  & \multicolumn{1}{c}{$\Rightarrow$} & \multicolumn{1}{c}{$\Leftarrow$}
  & \multicolumn{1}{c}{$\Rightarrow$} & \multicolumn{1}{c}{$\Leftarrow$} & \\
  \midrule
  \bf WMT20 Systems & & & & & & & \\
  ~~~~\citet{Shi:2020:WMT}       & 38.6 & 28.8 & ~~~- & ~~~- & ~~~- & ~~~- & - \\
  ~~~~\citet{Zhang:2020:WMT}     & 40.8 & ~~~- & 34.8 & 20.4 & ~~~- & ~~~- & - \\
  ~~~~\citet{Molchanov:2020:WMT} & ~~~- & ~~~- & ~~~- & ~~~- & 31.9 & 39.6 & - \\
  \midrule
  \bf Our Implemented Systems & & & & & & & \\
  ~~~~Baseline                   & 42.3 & 28.4 & 35.8 & 20.9 & 32.3 & 41.4 & 33.5 \\
  ~~~~Tag
    & \textbf{43.4}$^{\Uparrow}$ 
    & \underline{29.2}$^{\uparrow}$ 
    & \underline{36.3} 
    & \textbf{21.9}$^{\Uparrow}$ 
    & \underline{32.7} 
    & \textbf{42.5}$^{\Uparrow}$
    & \underline{34.3} \\
  ~~~~Tune
    & \underline{43.3}$^{\Uparrow}$
    & \textbf{29.7}$^{\Uparrow}$
    & \textbf{36.6}$^{\uparrow}$
    & \underline{21.8}$^{\uparrow}$
    & \textbf{32.9}$^{\uparrow}$
    & \underline{42.2}$^{\uparrow}$
    & \textbf{34.4} \\
  \bottomrule
  \end{tabular}
  \caption{Sacre BLEU reported on the WMT20 test sets. ``Tag" and ``Tune" denote the bias-tagging and fine-tuning, respectively. 
   We highlight the \textbf{highest} score in bold and the \underline{second-highest} score with underlines.
  ``$\uparrow$/$\Uparrow$'' denotes significantly better than the baseline with $p < 0.05$ and $p < 0.01$, respectively.
  For comparison, we list three systems that use vanilla Transformer models trained on the bilingual data in the WMT20 competition.}
  \label{tab:main}
\end{table*}

In Section~\ref{sec:obs-languagecoveragebias} we show that the target-original data performs poorly in translating content words due to the language coverage bias. 
Accordingly, simply using the full training data without distinguishing the original languages is sub-optimal for model training.
Based on these findings, we propose to address the language coverage bias by explicitly distinguishing between the source- and the target-original data (Section~\ref{sec:distinguish}).
We then investigate whether the performance improvement still holds in the monolingual data augmentation scenario (Section~\ref{sec:data-augmentation}), where the language coverage bias problem is more severe due to the newly introduced dataset in source or target language.

\subsection{Bilingual Data Utilization}
\label{sec:distinguish}
In this section, we aim to improve bilingual data utilization through explicitly distinguishing between the source- and target-original training data.
\paragraph{Methodology} We distinguish original languages with two simple and effective methods:
\begin{itemize}[leftmargin=*]
    \item {\em Bias-Tagging}: Tagging is a commonly-used approach to distinguishing between different types of examples, such as different languages ~\cite{Aharoni:2019:massively,Riley:2020:ACL} and synthetic vs authentic examples~\cite{Caswell:2019:WMT}. In this work, we attach a special tag to the source side of each target-original example, which enables NMT models to distinguish it from the source-original ones in training.
    \item {\em Fine-Tuning}: Fine-tuning~\cite{luong2015stanford} is a useful method to help knowledge transmit among data from different distributions. We pre-train NMT models on the full training data that consists of both the source- and target-original data, and then fine-tune them on only the source-original data. For fair comparison, the total training steps of the pre-training and fine-tuning stages are the same as the baseline.
\end{itemize}

\paragraph{Translation Performance}
Table~\ref{tab:main} depicts the results on the benchmarking datasets.
For comparison, we also list the results of several baselines using the vanilla Transformer architecture trained on the constrained bilingual data in the WMT20 competition~\cite{Barrault:2020:WMT}.
Clearly, both the bias tagging and fine-tuning approaches consistently improve translation performance on all benchmarks, which confirms our claim of the necessity of explicitly distinguishing target-original examples in model training.

\begin{table}[t]
  \centering
  \scalebox{0.89}{
  \begin{tabular}{l ccc ccc}
  \toprule
  \multirow{2}{*}{\bf Method}  &  \multicolumn{3}{c}{\bf En$\Rightarrow$Zh}  &  \multicolumn{3}{c}{\bf En$\Leftarrow$Zh}\\
  \cmidrule(lr){2-4} \cmidrule(lr){5-7}
       & \em noun & \em verb & \em adj  & \em noun & \em verb & \em adj\\
  \midrule
  Baseline & 70.7 & 61.0 & 67.9 & 60.2 & 43.6 & 61.4 \\
  Tag  & \bf 72.3 & \bf 62.3 & 67.9 & \underline{60.7} & \underline{43.9} & \bf 62.2 \\
  Tune & \underline{71.8} & \underline{61.9} & \bf 68.4 & \bf 61.1 & \bf 44.1 & \bf 62.2 \\
  \bottomrule
  \end{tabular}}
  \caption{F-measure of different types of content words on the WMT20 En$\Leftrightarrow$Zh test sets. Results on other languages can be found in Appendix (Table~\ref*{tab:main-adequacy-other-languages}).}
  \label{tab:main-adequacy}
\end{table}

\begin{table}[t]
  \centering
  \scalebox{0.9}{
  \begin{tabular}{l cc cc cc}
  \toprule
  {\bf Data} & \multicolumn{2}{c}{\bf En-Zh}  &  \multicolumn{2}{c}{\bf En-Ja} &  \multicolumn{2}{c}{\bf En-De}\\
  \cmidrule(lr){2-3} \cmidrule(lr){4-5} \cmidrule(lr){6-7}
  \bf Origin &   $\Rightarrow$   &   $\Leftarrow$    &   $\Rightarrow$   &   $\Leftarrow$    &   $\Rightarrow$   &   $\Leftarrow$\\
  \midrule
  Baseline & \textbf{13.7} & \textbf{13.4} & \textbf{17.4} & \textbf{15.4} & \textbf{16.3} & \textbf{17.8} \\
  Tag & \textbf{13.7} & \textbf{13.4} & \underline{17.5} & \textbf{15.4} & \underline{16.4} & \textbf{17.8} \\
  Tune & \underline{13.8} & \textbf{13.4} & \textbf{17.4} & \textbf{15.4} & \textbf{16.3} & \underline{17.9} \\
  \bottomrule
  \end{tabular}}
  \caption{PPL at the syntactic level on the test sets. We abstract the content words to rule out the language coverage bias when measuring fluency.}
  \label{tab:fluency-main}
\end{table}

\begin{table*}[t]
  \centering
  \begin{tabular}{cccc cccc cccc}
  \toprule
  \multirow{2}{*}{\bf \#}   &   \multicolumn{2}{c}{\bf Monolingual}  & \bf Bilingual & \multicolumn{4}{c}{\bf X$\Rightarrow$En}  & \multicolumn{4}{c}{\bf En$\Rightarrow$X} \\
  \cmidrule(lr){2-3} \cmidrule(lr){4-4} \cmidrule(lr){5-8} \cmidrule(lr){9-12}
  & \em Data &   \em Tagging  & \em Fine-Tune    &  \em Zh    &   \em Ja  &   \em De  &   \em Ave.   &  \em Zh    &   \em Ja  &   \em De  &   \em Ave.\\
  \midrule
  1 & \multirow{2}{*}{\texttimes}  &  \multirow{2}{*}{n/a} & \texttimes  &   28.4 &   20.9    &   41.4    &  30.2  &   42.3 & 35.8 &   32.3    &  36.8\\
  2 & & & \checkmark  &   \em 29.7 &   \em 21.8    &   \em 42.2   &  \em 31.2  &   \em 43.3    &   \em 36.6    &   \em 32.9 & \em 37.6 \\
  \midrule
  3 & \multirow{4}{*}{\checkmark}   &   \texttimes   &   \multirow{2}{*}{\texttimes} &  {\cellcolor{blue!22}} 28.9   &  {\cellcolor{blue!22}}  21.2  & {\cellcolor{blue!22}} 38.1  &   {\cellcolor{blue!22}} 29.4 & {\cellcolor{red!22}} \em 44.2  & {\cellcolor{red!22}} \em 36.8  & {\cellcolor{red!22}} \em 32.8  & {\cellcolor{red!22}} \em 37.9 \\
  4 & & \checkmark &  &  {\cellcolor{blue!22}}  \em 29.4    &  {\cellcolor{blue!22}} \em 21.2  & {\cellcolor{blue!22}} \em 41.5&   {\cellcolor{blue!22}} \em 30.7 & {\cellcolor{red!22}} 43.1  & {\cellcolor{red!22}} 36.4  & {\cellcolor{red!22}} 32.3  & {\cellcolor{red!22}} 37.3\\
  \cmidrule{3-12}
  5 & & \texttimes & \multirow{2}{*}{\checkmark} & {\cellcolor{blue!22}} 30.4 & {\cellcolor{blue!22}} 22.1 & {\cellcolor{blue!22}} 42.3 & {\cellcolor{blue!22}} 31.6 & {\cellcolor{red!22}} \bf 45.1 & {\cellcolor{red!22}} \bf 37.6 & {\cellcolor{red!22}} \bf 33.4 & {\cellcolor{red!22}} \bf 38.7 \\
  6 & &  \checkmark &  & {\cellcolor{blue!22}} \bf 30.6 & {\cellcolor{blue!22}} \bf 22.2 & {\cellcolor{blue!22}} \bf 42.7 & {\cellcolor{blue!22}} \bf 31.8 & {\cellcolor{red!22}} 44.7 & {\cellcolor{red!22}} 36.9 & {\cellcolor{red!22}} 33.3 & {\cellcolor{red!22}} 38.3 \\
  \bottomrule
  \end{tabular}
  \caption{Translation performance of augmenting English monolingual data with different strategies: {back-translation} for X$\Rightarrow$En tasks ({\color{blue!50} blue cells}), and {forward-translation} for En$\Rightarrow$X tasks ({\color{red!50} red cells}).
  {``Tagging" denotes adding a special tag to each synthetic sentence pair~\cite{Caswell:2019:WMT}.}
  ``Fine-Tune'' denotes fine-tuning the pre-trained NMT models on the source-original bilingual data, as described in Section~\ref{sec:distinguish}.}
  \label{tab:data-English-augmentation}
\end{table*}

\paragraph{Analysis}
Recent studies have shown that generating human-translation like texts as opposed to original texts can improve the BLEU score~\cite{Riley:2020:ACL}.
To validate that the improvement is partially from alleviating the content-dependent language coverage bias, we examine the translation adequacy of content words on the test sets, as listed in Table~\ref{tab:main-adequacy}.
The results indicate that explicitly distinguishing between the source- and target-original data improves the translation of content words (e.g., nouns), which is closely related to the language coverage bias problem.
Table~\ref{tab:fluency-main} lists the translation fluency at the syntactic level, where the proposed approaches maintain the syntactic fluency.

\subsection{Monolingual Data Augmentation}
\label{sec:data-augmentation}

In this section, we aim to provide some insights where monolingual data augmentation improves translation performance, and investigate whether our approach can further improve model performance in this scenario that potentially suffers more from the language coverage bias problem.

For fair comparison across language pairs, we augment NMT models with the same English monolingual corpus as described in Section~\ref{sec:setup}. We down-sample the large-scale monolingual corpus to the same amount as that of the bilingual corpus in each language pair, in order to rule out the effect of the scale of synthetic data~\cite{Edunov:2018:BTScale,Fadaee:2018:difficultwords}.
We use back-translation~\cite{Sennrich:2016:ACL} to augment the English monolingual data for the task of translating from another language to English (``X$\Rightarrow$En''), and use forward-translation for the task in the opposite translation direction (``En$\Rightarrow$X'').
Table~\ref{tab:data-English-augmentation} lists the results, where several observations can be made.

\paragraph{Explaining Data Augmentation with Language Coverage Bias}
Concerning the monolingual data augmentation methods (Rows 3-4), the vanilla back-translation (Row 3) harms the translation performance on average, while the vanilla forward-translation improves the performance, which is consistent with the findings in previous studies~\cite{Edunov:2020:ACL,Marie:2020:ACL}.
~\citet{Caswell:2019:WMT} have shown that the tagging strategy works for back-translation while fails for forward-translation, and our results confirm these findings. 
Both phenomena can be attributed in part to the language coverage bias problem. Back-translated data originates from the target language, and thus suffers more from the language coverage bias problem. Accordingly, directly using the back-translated data is sub-optimal, while tagged back-translation recovers translation performance by distinguishing training examples with different origins, which is consistent with our results in Table~\ref{tab:main}. 
In contrast, the language coverage bias problem does not exist for source-side monolingual data (i.e. the same original language). Therefore, the vanilla forward-translation can improve translation performance, while tagged forward-translation performs worse.

\paragraph{Improving Data Augmentation}
Our approach (Row 2) achieves comparable improvements of translation performance with the monolingual data augmentation approaches (e.g. averaged BLEU: 31.2 vs. 30.7, and 37.6 vs. 37.9), while we do not use additional monolingual data to train the models.\footnote{The monolingual data is only used to detect the original languages of training data and is invisible in model training.}
Combining them can further improve performance (Rows 5-6), indicating that the two types of approaches are complementary to each other. This is straightforward, since our approach better exploits the bilingual data, while data augmentation introduces new knowledge from additional monolingual data.
In addition, our approach consistently improves performance over both vanilla and tagged augmentation approaches, making it more robust in practical application across datasets.

\section{Related Work}

Our work is inspired by three lines of research in the NMT community.

\subsection{Translationese}
Recently, the effect of translationese in NMT evaluation has attracted increasing attention~\cite{Zhang:2019:WMT,Bogoychev:2019:domain,Edunov:2020:ACL,Graham:2020:EMNLP}. 
\citet{Graham:2020:EMNLP} show that the source-side translationese texts can potentially lead to distortions in automatic and human evaluations. Accordingly, the WMT competition starts to use only source-original test sets for most translation directions since 2019. 
Our study reconfirms the necessity of distinguishing the source- and target-original examples and takes one step further to distinguish examples in training data.
Complementary to previous works, we investigate the effect of language coverage bias on machine translation, which is related to the content bias rather than the language style difference.
\citet{Shen:2021:Domain-Mismatch} also reveal the context mismatch between texts from different original languages. To alleviate this problem, they proposed to combine back- and forward-translation by introducing additional monolingual data, while we focus on better exploiting bilingual data by distinguishing the original languages, which is also helpful for back- and forward-translation.

\citet{Lembersky:2011:EMNLP,Lembersky:2012:EACL} propose to adapt machine translation systems to generate texts that are more similar to human-translations, while \citet{Riley:2020:ACL} propose to model human-translated texts and original texts as separate languages in a multilingual model and perform zero-shot translation between original texts. \citet{Riley:2020:ACL} and our work both aim to better utilize the bilingual training data. They aim to guide NMT models to produce original text, while we focus on improving translation adequacy by alleviating the language coverage bias problem.

\subsection{Data Augmentation}
Concerning model training, recent works find that back-translation can harm the translation of source-original test set, and attribute the quality drop to the stylistic and content-independent differences between translationese and original texts~\cite{Edunov:2020:ACL,Marie:2020:ACL}.
In this work, we empirically show that language coverage bias is another reason for the performance drop of back-translation, as well as the different performances between tagged forward-translation and tagged back-translation~\cite{Caswell:2019:WMT}.
In addition, we show that our approach is also beneficial for data augmentation approaches, which can further improve the translation performance over both back-translation and forward-translation.

\subsection{Domain Adaptation}
Since high-quality and domain-specific parallel data is usually scarce or even unavailable, domain adaptation approaches are generally employed for translation in low-resource domains by leveraging out-of-domain data~\cite{Chu:2018:survey}.
Languages can be also regarded as different domains, since articles in different languages cover different topics~\cite{Bogoychev:2019:domain}. Starting from this intuition, we distinguish examples with different original languages with tagging~\cite{Aharoni:2019:massively} and fine-tuning~\cite{luong2015stanford}, which are commonly-used in domain adaptation and multi-lingual translation tasks.

Our work also benefits domain adaptation: distinguishing original languages in general domain data consistently improves translation performance of NMT models in several specific domains (Table~\ref*{tab:domain} in Appendix), making these models better start points for further domain adaptation.

\section{Conclusion and Future Work}

In this work, we first systematically examine why the language coverage bias problem is important for NMT models. We conducted extensive experiments on six WMT20 translation benchmarks. Empirically, we find that source-original data and target-original data differ significantly at the text content, and using target-original data together without discrimination is sub-optimal.
Based on these observations, we propose two simple and effective approaches to distinguish the source- and target-original training data, which obtain consistent improvements in all benchmarks.

Furthermore, we link language coverage bias to two well-known problems in monolingual data augmentation, namely the performance drop of back-translation, and the different behaviors between tagged back-translation and tagged forward-translation.
We show that language coverage bias can be considered as another reason for these problems, and fine-tuning on the source-original bilingual training data can further improve performance over both back- and forward-translation.

Future directions include exploring advanced methods to better alleviate the language coverage bias problem, as well as validating on other language pairs.
It is also interesting to investigate the language coverage bias problem in multilingual translation, where we can better understand this problem by considering language family.

\section*{Acknowledgments}
We thank all anonymous reviewers for their insightful comments and suggestions for this work. We also sincerely thank Cunxiao Du and Wenxiang Jiao for their valuable help and advice.
This work was supported by the National Key R\&D Program of China (No. 2017YFB0202204), National Natural Science Foundation of China (No. 61925601, No. 61772302), and the Tencent AI Lab Rhino-Bird Focused Research Program.

\bibliographystyle{acl_natbib}
\bibliography{anthology,acl2021}

\clearpage

\appendix

\section{Appendices}

\subsection{Data Preprocessing}
\label{sec:appendix-data-process}
We used all the parallel corpora provided by WMT20 and filtered sentences that are longer than 250 words.
We tokenized English and German sentences with Moses~\cite{Koehn:2007:moses}, and segmented Chinese and Japanese sentences with Jieba\footnote{https://github.com/fxsjy/jieba} and Mecab\footnote{https://taku910.github.io/mecab} respectively. 
We employed Byte pair encoding (BPE) \cite{Sennrich:2016:bpe} with 32K merge operations for all language pairs. 
Specifically, we jointly trained the BPE code on both sides in En$\Leftrightarrow$De and independently learned the BPE code on each side in En$\Leftrightarrow$Zh and En$\Leftrightarrow$Ja. 

As for the monolingual data, we combined the newscrawl data from 2017 to 2019 for English and German. Since the newscrawl corpora for Chinese and Japanese are significantly smaller, we augmented these two languages with the commoncrawl corpus. We preprocessed the monolingual data with the same rules as parallel data.
Finally, we randomly selected 41.0M sentences for each language (i.e., En, De, Zh, Ja), which were used to train the language detection models. For data augmentation, to rule out the effect of the ratio between synthetic and authentic data, we down-sampled the monolingual data to the same amount as the bilingual data for each language pair.

We used spaCy\footnote{https://spacy.io} to perform the Part-Of-Speech (POS) tagging for each language. Nouns, verbs, and adjectives belong to content words and the others belong to function words.

\subsection{More Details of Model Training}
\label{sec:appendix-model-train}
In this work, we generally followed the default hyper-parameters used in~\citet{Vaswani:2017:NeurIPS} except the batch size. Recent studies showed that training on large batches can further boost model performance~\cite{Ott:2018:WMT,Wu:2018:ICLR}. Accordingly, we followed them to train models with batches of approximately 460k tokens, using Adam~\citep{kingma2015adam} with $\beta_{1}=0.9$, $\beta_{2}=0.98$ and $\epsilon=10^{-8}$. 
We used the same cosine learning rate schedule as \citet{Wu:2018:ICLR}, where the learning rate was warmed up linearly in the first 10K steps, and then decayed following a cosine rate within a single cycle. By default, NMT models and language models were both trained for 30k steps with the aforementioned batch size. Each model was trained using 8 NVIDIA V100 GPUs for about 20 hours.

\subsection{Effect of Detection Methods on Translation Performance}
To further compare our proposed original language detection method and the FT classifier~\cite{Riley:2020:ACL}, we fine-tune the NMT model pre-trained on the whole training set using the source-original data detected by the two methods. Note that the two detection methods are developed using the same monolingual data sets. For fair comparison, the fine-tuning sets are of the same amount (50\% of the whole training set) between the two methods in this experiment. Table~\ref{tab:effect-of-detection} lists the results, indicating that our method performs better in detecting original languages in large-scale parallel data.

\begin{table}[h]
    \centering
    \begin{tabular}{cc}
    \toprule
    Fine-Tune Data & BLEU \\
    \midrule
    \texttimes  & 27.5 \\
    \midrule
    FT          & 27.8 \\
    Ours        & \bf 28.4 \\
    \bottomrule
    \end{tabular}
    \caption{Effect of original language detection methods. The results are reported on the validation set of the Zh$\Rightarrow$En translation task.}
    \label{tab:effect-of-detection}
\end{table}

\begin{table*}[p]
  \centering
  \begin{tabular}{crrrrrrrrr}
  \toprule
  \multirow{2}{*}{\bf Data}  &  \multicolumn{3}{c}{\bf 10\%} & \multicolumn{3}{c}{\bf 30\%} & \multicolumn{3}{c}{\bf 50\%} \\
  \cmidrule(lr){2-4} \cmidrule(lr){5-7} \cmidrule(lr){8-10}
    & \em All & \em Content & \em Function  & \em All & \em Content & \em Function  & \em All & \em Content & \em Function\\
  \midrule
  Random & 20 & 51 & 0  & 7 & 18 & 0 & 4 & 10 & 0\\
  S vs T & 2660 & 5096 & 961 & 1371 & 2731 & 486 & 745 & 1503 & 261\\
  \bottomrule
  \end{tabular}
  \caption{JS divergence ($\times 10^{-5}$) of the vocabulary distributions between the source- and target-original training data (``S vs T'') for different labeled ratios on En$\Leftrightarrow$Zh. For reference, we also report the JS divergence between two sets of randomly selected examples (``Random'', non-overlap).}
  \label{tab:language-bias-ratios}
\end{table*}

\begin{table*}[p]
  \centering
  \begin{tabular}{crrrrrrrrr}
  \toprule
  \multirow{2}{*}{\bf Data}  &  \multicolumn{3}{c}{\bf En-Zh} & \multicolumn{3}{c}{\bf En-Ja} & \multicolumn{3}{c}{\bf En-De} \\
  \cmidrule(lr){2-4} \cmidrule(lr){5-7} \cmidrule(lr){8-10}
  & \em All & \em Content & \em Function  & \em All & \em Content & \em Function  & \em All & \em Content & \em Function\\
  \midrule
  Random & 4 & 10 & 0 & 8 & 17 & 0 & 2 & 4 & 0 \\
  S vs T & 745 & 1503 & 261 & 1687 & 2910 & 666 & 870 & 1622 & 250  \\
  \bottomrule
  \end{tabular}
  \caption{JS divergence ($\times 10^{-5}$) of the vocabulary distributions between the source- and target-original training data for different language pairs. 50\% examples are treated as source-original and the others are treated as target-original. For reference, we also report the JS divergence between randomly selected 50\% examples and the others (``Random'', non-overlap).}
  \label{tab:language-bias-lanuages}
\end{table*}

\begin{table*}[p]
  \centering
  \begin{tabular}{c ccc ccc ccc ccc}
  \toprule
  {\bf Data}  &  \multicolumn{3}{c}{\bf En$\Rightarrow$Ja}  &  \multicolumn{3}{c}{\bf En$\Leftarrow$Ja}  &  \multicolumn{3}{c}{\bf En$\Rightarrow$De}  &  \multicolumn{3}{c}{\bf En$\Leftarrow$De}\\
  \cmidrule(lr){2-4} \cmidrule(lr){5-7} \cmidrule(lr){8-10} \cmidrule(lr){11-13}
  \bf Origin & \em noun & \em verb & \em adj  & \em noun & \em verb & \em adj   & \em noun & \em verb & \em adj  & \em noun & \em verb & \em adj\\
  \midrule
  Target & 60.9 & 47.7 & 62.1 & 44.5 & 29.8 & 46.2 & 70.5 & 54.3 & 58.4 & 70.8 & 53.6 & 67.0 \\
  Source & \bf 61.4 & \bf 51.8 & \bf 63.5 & \underline{49.5} & \underline{31.8} & \underline{50.1} & \underline{72.1} & \underline{55.0} & \underline{60.3} & \bf 75.3 & \underline{55.3} & \underline{70.2} \\
  Both   & \underline{61.3} & \underline{51.7} & \underline{63.2} & \bf 50.7 & \bf 32.1 & \bf 50.4 & \bf 72.7 & \bf 56.6 & \bf 60.4 & \underline{74.9} & \bf 55.7 & \bf 71.0 \\
  \bottomrule
  \end{tabular}
  \caption{Translation adequacy of different types of content words measured by F-measure~\cite{Neubig:2019:compare}. The results are reported on the validation sets.}
  \label{tab:adequacy-other-languages}
\end{table*}

\begin{table*}[p]
  \centering
  \begin{tabular}{l ccc ccc ccc ccc}
  \toprule
  \multirow{2}{*}{\bf Method}  &  \multicolumn{3}{c}{\bf En$\Rightarrow$Ja}  &  \multicolumn{3}{c}{\bf En$\Leftarrow$Ja}  &  \multicolumn{3}{c}{\bf En$\Rightarrow$De}  &  \multicolumn{3}{c}{\bf En$\Leftarrow$De}\\
  \cmidrule(lr){2-4} \cmidrule(lr){5-7} \cmidrule(lr){8-10} \cmidrule(lr){11-13}
       & \em noun & \em verb & \em adj  & \em noun & \em verb & \em adj   & \em noun & \em verb & \em adj  & \em noun & \em verb & \em adj\\
  \midrule
  Baseline & 62.0 & 53.0 & 59.1 & 54.0 & 35.5 & 50.4 & 66.7 & 48.1 & 53.6 & 75.0 & 54.0 & 70.2 \\
  Tag & \underline{62.5} & \underline{53.3} & \underline{61.1} & \bf 55.7 & \underline{36.5} & \bf 52.4 & \bf 67.1 & \underline{48.6} & \bf 54.0 & \bf 76.1 & \underline{54.4} & \underline{70.7} \\
  Tune & \bf 62.8 & \bf 53.7 & \bf 61.7 & \underline{55.3} & \bf 36.9 & \underline{51.8} & \bf 67.1 & \bf 48.7 & \bf 54.0 & \underline{75.7} & \bf 54.8 & \bf 70.8 \\
  \bottomrule
  \end{tabular}
  \caption{Translation adequacy of different types of content words measured by F-measure~\cite{Neubig:2019:compare}. The results are reported on the test sets.}
  \label{tab:main-adequacy-other-languages}
\end{table*}

\begin{table}[ht]
  \centering
  \begin{tabular}{l cc}
  \toprule
  \bf Domain & \bf Baseline & \bf Ours \\
  \midrule
  Business & 40.4 & \bf 40.8 \\
  Crime & 34.8 & \bf 35.5 \\
  Entertainment & 28.8 & \bf 30.0 \\
  Politics & 39.5 & \bf 40.3 \\
  Sci-Tech & 38.2 & \bf 39.9 \\
  Sport & \bf 31.5 & \bf 31.5 \\
  World & 38.8 & \bf 38.9 \\
  \midrule
  Overall & 36.6 & \bf 37.2 \\
  \bottomrule
  \end{tabular}
  \caption{Transformer performance on the validation set of the En$\Rightarrow$Zh task. We split the whole validation set into several parts by the domain tag. ``Ours" denotes the ``Bias-Tagging" approach as described in Section~\ref{sec:distinguish}. The results indicate that distinguishing data with different original languages in the general domain training data can improve the performance of NMT models in many specific domains, making the models better start points for further domain adaptation.}
  \label{tab:domain}
\end{table}

\subsection{Divergence of Vocabulary Distributions}
\label{sec:importance-languages}
In this section, we report the JS divergence of the vocabulary distributions in more cases. Table~\ref{tab:language-bias-ratios} lists the results for different ratios $R\%$ on En$\Leftrightarrow$Zh, and Table~\ref{tab:language-bias-lanuages} shows the results on all language pairs. The results show that the divergence of vocabulary distributions between the source- and target-original data is substantially larger than that between randomly split data, which reconfirms the existence of language coverage bias.

\subsection{Effect of Language Coverage Bias for Other Language Pairs}
Table~\ref{tab:adequacy-other-languages} lists the translation adequacy of NMT models trained on only the source- or target-original data and on both of them. The results are reported on En$\Leftrightarrow$De and En$\Leftrightarrow$Ja, which exhibit the same trend as that on En$\Leftrightarrow$Zh (Table~\ref{tab:adequacy} in the main paper), indicating that the target-original data performs poorly in translating content words.

\subsection{Translation Adequacy on Test Sets for Other Language Pairs}
We report the translation adequacy on test sets for En$\Leftrightarrow$De and En$\Leftrightarrow$Ja in Table~\ref{tab:main-adequacy-other-languages}, corresponding to Table~\ref{tab:main-adequacy} in the main paper. The results show that explicitly distinguishing the source- and target-original training data can consistently improve the translation adequacy for content words on all the six translation tasks.

\subsection{Translation Performance in Specific Domains}
We evaluate NMT models trained with and without explicit distinguishing between the source- and target-original data in several specific domains. The results are shown in Table~\ref{tab:domain}, suggesting that our method can improve the translation performance of NMT models in several specific domains, which can be combined with further domain adaptation approaches.

\end{document}